\documentclass[letterpaper, 10 pt, conference]{ieeeconf}  % Comment this line out if you need a4paper

\IEEEoverridecommandlockouts                              % This command is only needed if 
\usepackage{graphics} % for pdf, bitmapped graphics files
\usepackage{epsfig} % for postscript graphics files
\usepackage{mathptmx} % assumes new font selection scheme installed
\usepackage{times} % assumes new font selection scheme installed
\usepackage{amsmath} % assumes amsmath package installed
\usepackage{amssymb}  % assumes amsmath package installed
\usepackage{hyperref}
\usepackage{booktabs}
 \usepackage{graphicx,tabularx}
 \usepackage{svg}

\usepackage{subcaption}

\usepackage{xcolor}

\title{\LARGE \bf
Local Gaussian Modifiers (LGMs): UAV dynamic trajectory generation for onboard computation
}
\author{Miguel Fernandez-Cortizas$^{1}$, David Perez-Saura$^{1}$, Javier Rodriguez-Vazquez$^{1,2}$, Pascual Campoy$^{1}$% <-this % stops a space
\thanks{ $^{1}$Computer Vision and Aerial Robotics group (CVAR), Centre for Automation and Robotics (CAR), Universidad Politécnica de Madrid (UPM),  Madrid, Spain. 
E-mail: miguel.fernandez.cortizas@upm.es }%
\thanks{$^{2}$Department of Artificial Intelligence, Universidad Polit\'ecnica de Madrid,   Madrid, Spain}
}

\begin{document}

\maketitle
\thispagestyle{empty}
\pagestyle{empty}

%%%%%%%%%%%%%%%%%%%%%%%%%%%%%%%%%%%%%%%%%%%%%%%%%%%%%%%%%%%%%%%%%%%%%%%%%%%%%%%%
\color{black}
\begin{abstract}

% Agile drones are important for these reasons. trajectory generation also. Nowadays it is doing static and off board. dynamic and on board presents difficulties. in this paper we present a new approach good for dynamic trajectory generation and on board computing. we have test our approach both in simulation and real experiments. we have achieve this. 

Agile autonomous drones are becoming increasingly popular in research due to the challenges they represent in fields like control, state estimation, or perception at high speeds. When all algorithms are computed onboard the uav, the computational limitations make the task of agile and robust flight even more difficult. 
One of the most computationally expensive tasks in agile flight is the generation of optimal trajectories that tackles the problem of planning a minimum time trajectory for a quadrotor over a sequence of specified waypoints. When these trajectories must be updated online due to changes in the environment or uncertainties, this high computational cost can leverage to not reach the desired waypoints or even crash in cluttered environments. 
In this paper, a fast lightweight dynamic trajectory modification approach is presented to allow modifying computational heavy trajectories using Local Gaussian Modifiers (LGMs), when recalculating a trajectory is not possible due to the time of computation.

Our approach was validated in simulation, being able to pass through a race circuit with dynamic gates with top speeds up to 16.0 m/s, and was also validated in real flight reaching speeds up to 4.0 m/s in a fully autonomous onboard computing condition.

\end{abstract}
%%%%%%%%%%%%%%%%%%%%%%%%%%%%%%%%%%%%%%%%%%%%%%%%%%%%%%%%%%%%%%%%%%%%%%%%%%%%%%%%
\section*{SUPLEMENTARY MATERIAL}
Video of the experiments: \url{https://vimeo.com/683638197}.
Released code : \url{https://github.com/miferco97/dynamic_trajectory_generator}

\color{black}
\section{INTRODUCTION}
% Talk about the use of autonomous tasks in different uses 
Multirotors are highly versatile and agile aerial robotic platforms, thanks to their maneuverability and simplicity. Nowadays, these vehicles are being used in several tasks such as inspection, delivery, cinematography, or search-and-rescue \cite{romero2021model}. Nowadays, most drone applications require a human pilot who is in charge of controlling them. The research community and the industry are working to achieve a higher level of autonomy in drones, which will leverage to perform complex tasks without needing human intervention.

Drones can carry on-board computers, which allows the drone to perform complex tasks like interpret the environment, generate a map, or compute complex trajectories, without relying on stable and fast communication between the aircraft and the ground, what improves the robustness of the system. However, due to the limited weight that a drone is capable of carrying, the power of these computers is limited.

Autonomous drone flight needs various components working together in a coordinate way, such as state estimation, control, environment perception, or planning components \cite{imav2021-5:m_fernandezcortizas_et_al}.  When all this components are wanted to be computed onboard the drone in real time, computing resources become an important limitation. %\tb{makes harder making drones fly fast, due to the reaction times that this implies ??}.

When UAVs fly in the real world, they have to deal with lots of uncertainties in self-localization, environment recognition, and dynamic modeling, which often are combined with changes in the environment, this is why being able to adapt to different conditions is fundamental.

For the control modules, the most popular strategies, like geometric controllers \cite{lee2010geometric}, quaternion-based controllers \cite{fresk2013full} or Model Predictive controllers (MPC) \cite{bangura2014real}, relies on a previously computed dynamic feasible sequence of states and inputs to track. The problem of generating this sequence is called trajectory generation.

Generating trajectories that change continuously can be a computational expensive task, involving an important amount of time. During the time spent modifying the trajectory, the drone can be flying, following the previous trajectory. If the time for the trajectory generation is too long, it can result in a collision because the new trajectory is computed too late. When all computations are done on a small computer onboard the drone, the calculation time increases. 

All these things make it necessary to have a computational cheap and fast way to modify the trajectory when there is not time enough to generate a new trajectory. In this work, we focus on developing a fast and lightweight dynamic trajectory generation that is able to adapt to environmental changes as long as the drone is flying through it.

\subsection{Related work}

% Current research is pushing forward to make agile autonomous drones that can fly through an environment as fast as possible. This task is easily translated into passing through a set of waypoints like in a drone race. The problem of generating a trajectory that passes through multiple waypoints in minimal time, it is known as time-optimal planning. 

The formulation of the trajectory planning problem for multirotors has evolved from the simple shortest path approach to complex minimum time optimization problems. 
For simple point-mass systems, time optimal trajectories can be computed in closed-form, resulting in bang-bang acceleration trajectories, which can easily be sampled over multiple waypoints. However,   multirotors are under-actuated systems \cite{bouabdallah2007full} \cite{mahony2012multirotor}, which means that there is a coupling between linear and rotational accelerations. This coupling causes problems at the moment of generating time optimal trajectories \cite{penin2017vision}. 

There are two main approaches for trajectory generation of drones. On the one hand, polynomial trajectory planning \cite{mellinger2011minimum}\cite{richter2016polynomial}, which is efficient computationally and exploits differentially flat output states, but the smoothness of the polynomials cannot take advantage of the full actuator potential of the aircraft.
On the other hand, there are discretized state space formulation approaches that uses nonlinear optimization to plan in a time-discretized state space using a more complex quadrotor model by taking advantage of the full quadrotor dynamics, such as the Complementary Progress Constraints (CPC) trajectory generator \cite{foehn2021time}. This approaches are computationally demanding, taking on the order of minutes or even hours to generate a trajectory. Due to this high computational cost, these trajectories are precomputed offline for a fixed and invariant circuit. 

Alternatively, there are other approaches that try to solve the problem of control and trajectory generation simultaneously, such as Model Predictive Contouring Control (MPCC) techniques \cite{romero2021model}. 

% Currently, there is a lot of research in this field, using MPC or neural networks to obtain an accurate model of the drone and achieve the best time-optimal trajectory thanks to the knowledge of the platform limits.
In real world applications, we have to deal with uncertainties and with changing or unknown environments in real time. That means we will have changes in the path that generate the necessity of changing the trajectory during the fly.

% Generating trajectories for autonomous vehicles in real scenarios is a long-known problem \cite{werling2010optimal}.
% As drones are sub-actuated system, some problems related to the trajectory generator has been studied \cite{penin2017vision}.
% differential flatness and B-Splines to parametrize the system trajectories in terms of a finite number of control points, which can then be optimized by Sequential Quadratic Programming /cite{}.

% In this thesis we exploit numerical optimization to design feasible trajectories satisfying several state, input and visual nonlinear constraints. With the help of differential flatness and B-spline parametrization we will propose an efficient replanning strategy inspired form Model Predictive Control to generate smooth and agile trajectories.
% One approach is to model the dynamics of the drone...

% Alternatively, some techniques decided to not model this and use differencial flatness representation of the quadrotor . "eliminates the need for computationally intensive sampling and simulation in the high dimensional state space of the vehicle during motion planning" \cite{richter2016polynomial}.

% we use ethz mav trajectory generator is based on \cite{richter2016polynomial} and optimized  \cite{burri2015real-time}.

\subsection{Contribution} 

In this work, we present a fast and lightweight methodology for generating adaptative trajectories that react to changes in the waypoints set in a smooth and agile way. Our approach consists of combining a polynomial trajectory generator for generating an optimal trajectory in snap, which constitutes the base trajectory (baseline), with local Gaussian modifiers (LGMs) that modify the baseline trajectory when recomputing this baseline trajectory is not feasible. Moreover, we present a strategy for stitching two polynomial trajectories in a smooth way. 
Compared to other trajectory generators, this one presents an organic approach in which it is taken into account that the drone is sampling his trajectory to generate smooth trajectories in the simplest and most transparent way for the user. 

% \color{-red!50}
% Una herramienta, método, para realizar modificaciones sobre una trayectoria optima en un tiempo muy reducido y con bajo coste computacional, ademas de permitir volver a la trayectoria optima tras esquivar el obstaculo. 
% trayectoria polinomica
% el codigo no es ROS, es C++. optimo, pequeño, para ponerlo en drones mas pequeños, aumentar el tiempo de reacción
%%%%%%%%%%%%%%%%%%%%%%%%%%%%%%%%%%%%%%%%%%%%%%%%%%%%%%%%%%%%%%%%%%%%%%%%%%%%%%%%
\color{black}
% \newpage
% \newpage

% \section{DYNAMIC TRAJECTORY GENERATION}
\section{METHODOLOGY}

% el modelo del uav es plano diferencialmente, 
% breve explicación
% ecuaciones
% alguna imagen quizas (o no, que se bsquen el paper)

% minimun snap trajectories
% constrains

% Local modifiers 
% como se generan las gaussianas. pq gaussianas. continuaidad en velocidad y aceleración.
% grafica de la generación en posición, velocidad y aceleración.
 
% Security zone -> dynamic minimum snap  trajectory regeneration 

% Stitching

% \subsection{Problem statement}

% Given a 3D occupancy map of an environment, we wish to efficiently compute feasible, minimum-snap trajectories that follow the shortest collision-free path from start to goal utilizing the full dynamic capabilities of the quadrotor
% Autonomous drone racing considers the problem of traversing a path in minimum time while passing through the designated waypoints (gates). To this end, polynomial-based multiwaypoint trajectory planning algorithms have been widely used in the literature due to their simplicity

% Quadrotor trajectory planning has been extensively studied in the literature. Polynomial trajectories are the first category of planning algorithms, such as the now widely used minimum-snap trajectories [30, 3]. Given a 4th order smooth trajectory, the full state at each point on the trajectory can be derived using the differential flatness property of the quadrotor

\subsection{Notation}
In this paper, we use the global frame $W$ to plan and generate all trajectories. For vectorial variables, functions, and constants, we use bold letters, as $\textbf{x}$. Tilde notation represents the new update of a variable, before been taken into the trajectory, e.g. $\tilde{\textbf{w}}$ means the waypoint position update before recalculating the new trajectory.

%\subsection{Methodology}
\subsection{Problem formulation}

% With this definition, we can formulate our problem as follows:

% We have a path of N dynamic wapoints that we want to traverse as optimal as posible with onboard computation.

% 
% side this threshold,  we are flying  blindly, so we are not able to correct the trajectory to ensure the waypoint traverse
% our goal is to develop a cheap trajectory modification method that allows u to keep recalculating (suboptimal) trajectories to the very last moment, increasing the success rate of reaching those waypoints

% I would also add that we assume that we are not reaching the UAV dynamics limits, so we can afford to doesnt take in account those, allowing us to leverage the computational constrains even more

% At lst, measurement uncertainty can be equivalent to dynamic waypoints, in the sense that when you are reaching the waypoint, the waypoint "moves" as you lower that uncertaint

Given a set of $N$ dynamic waypoints, we aim to compute agile dynamic trajectories $\mathbf{F}(t)$ that traverse through each waypoint as optimal as possible with onboard computation limitations, being able to modify the position of the dynamic waypoints as the quadrotor runs it. 

Due to computational constraints,  when reaching a waypoint, there is a temporal threshold $T_{security}$ from when we do not have enough time to recompute the trajectory using conventional methods.

Inside this threshold, we are flying blindly, so we are not able to correct the trajectory to ensure the waypoint traverse. Our goal is to develop a cheap trajectory modification method that allows us to keep recalculating trajectories to the very last moment, increasing the success rate of reaching those waypoints, although these modifications can lead to follow a suboptimal trajectory.

We define a dynamic waypoint $\mathbf{w} = [x,y,z]^t$ as a 3D point with an ID, whose position can change over time. Each waypoint may have other restrictions such as the velocity $\mathbf{\dot w}$ or acceleration $\mathbf{\ddot w}$ that the aircraft must have when passing through it.

% Mathematically, we denote a trajectory as a function $F(t):\mathbb{R}^+ \xrightarrow{} \mathbb{R}^3 \; ; F(t) \in C^3 $ where:
% \begin{align}
%     F(t) &= \mathbf{x}(t)  &F(t) \in \mathbb{R}^3\\
%     \dot{F}(t) &= \mathbf{\dot{x}}(t) = \mathbf{v}(t)  &\dot F(t) \in \mathbb{R}^3 \\
%     \ddot{F}(t) &= \mathbf{\ddot{x}}(t) = \mathbf{a}(t) &\ddot F(t) \in \mathbb{R}^3
% \end{align} 
% where $\mathbf{x}(t)=\{x(t),y(t),z(t)\}$ represents a position depending on time in the world coordinates and $\mathbf{\dot x}, \mathbf{\ddot x}$ its time derivatives.
For generating a base trajectory  $\mathbf{P}(t)$ we rely on a polynomial trajectory generator based on the Ritcher et al. \cite{richter2016polynomial} work. This approach can be used with other trajectory generators more sophisticated, but in this work we decided to use a simple one with a good trade between performance and computational cost. In this work, they generate piecewise polynomial minimum snap trajectories based on the differential flatness property of the quadrotor dynamics. This trajectory is expressed as: 
\begin{equation}
    \mathbf{P}(t)= \begin{cases}
    \;\sum_{i=0}^{n} c_{i,1} \; t^i \qquad t_0\le t < t_1 \\
    \;\sum_{i=0}^{n} c_{i,2} \; t^i \qquad t_1\le t < t_1 \\
    \;\qquad \vdots \\
    \;\sum_{i=0}^{n} c_{i,N} \; t^i \qquad t_{N-2}\le t < t_{N-1} 
    \end{cases}
\end{equation}
where $N$ represents the number of waypoints, $n$ the order of the polynomial, and $c_{i,j} \;;\;  i=0,..,n \;;\; j = 1,...,N$ the coefficients of each polynomial. More details about how to compute these trajectories can be found in \cite{richter2016polynomial}\cite{burri2015real-time}.

In our approach, we compute an adaptative trajectory that combines a polynomial trajectory $\mathbf{P}(t)$ with Local Gaussian Modifiers (LGMs). We define a $\mathbf{LGM}(t) : \mathbb{R}^+ \xrightarrow{} \mathbb{R}^3$ as:
\begin{equation}
    \mathbf{LGM}(t) = \mathbf{A} e^{ - \frac{(t - \mu)^2}{2\sigma^2}}
\end{equation}
where $\mathbf{A \in \mathbb{R}^3}$ represents the magnitude of the modification in the position of the waypoint on each axis and $\mu ,\sigma \in \mathbb{R}^+$ are constants that are computed when each LGM is created.
Each dynamic waypoint can have multiple LGMs associated with them.

% Due that each waypoint is tridimensional, the waypoint modifications will consist of modifications in three axes, we denote each set of three LGMs as $M(t)=[LGM_x(t),LGM_y(t),LGM_z(t)]^t$

With these components, we define the time evaluation of our dynamic trajectory $\mathbf{F(t)}$ as:
\begin{equation}
    \mathbf{F(t)} = \mathbf{P(t)} + \sum_{i=0}^{N_{w}} \sum_{j=0}^{N_{m_i}}\mathbf{LGM}_{i,j}(t)
\end{equation}
Where $N_w$ represents the number of waypoints of the trajectory and $N_{m_i}$ the number of modifications of the i-nth waypoint. 

In this work, we assume that we are not reaching the UAV dynamics limits when polynomial trajectories are computed, so we can afford to not take into account the limits in speed and acceleration when LGMs are taken into account, allowing us to leverage the computational constraints even more.

\subsection{Dynamic Trajectory Generation}
When an UAV is flying at high speeds, and the waypoint position changes, fast, reactiveness is fundamental for avoiding collisions . This reactivity is limited by the computational cost involved in generating these trajectories. When these trajectories are computed on onboard computational systems, this effect is even more notorious.

In this approach, we consider different ways to modify a trajectory depending on how fast a new trajectory can be generated in a safe way, ensuring that the trajectory generated will pass through this modified waypoint. 

For this task, we define a security time $T_{security}$:
\begin{equation}
    T_{security} = C_{security} \cdot T_{computation}(n);
\end{equation}

where $ T_{computation}(n) $ is a estimation of how much time a n-waypoints trajectory needs for being calculated, and $C_{security}$ a security constant for ensure that the UAV will have enough time for reacting after the trajectory were modified, in this work we use $C_{security} = 5$. 

% For estimating $T_{computation}(n)$, the trajectory generation time is measured online, as long as each trajectory is generated.
The estimation of $T_{computation}(n)$, is calculated online, based on the average time that previous trajectories of $n$ waypoints took to be calculated. This estimation is updated as long as new trajectories are calculated, taking into account the computational load of the the onboard computing during the flight. 

\begin{figure}[thpb]
      \centering
    %   \framebox{\parbox{3in}{We suggest that you use a text box to insert a graphic (which is ideally a 300 dpi TIFF or EPS file, with all fonts embedded) because, in an document, this method is somewhat more stable than directly inserting a picture.}}
      %\includegraphics[scale=1.0]{figurefile}
    %   \includegraphics[width=0.4\textwidth]{images/sec_time_d.jpg}
      \includegraphics[width=0.4\textwidth]{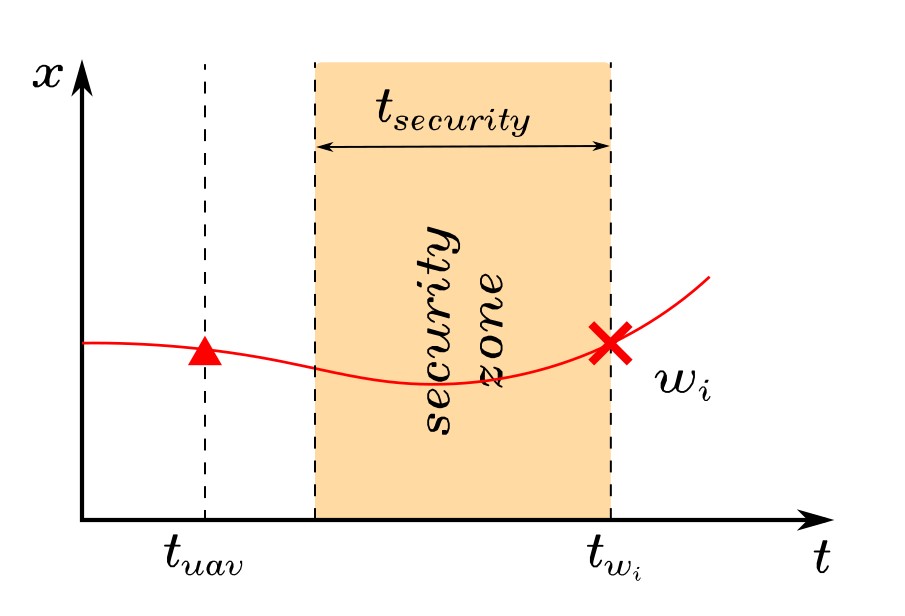}
      \caption{Diagram representing the security zone based in $t_{security}$ respect the next waypoint $w_i$, when the UAV is at $t_{uav}$ }
      \label{sec_zone}
   \end{figure}

With this $T_{security}$ we can define a security zone $SZ$ as the period of time where the UAV is less time away from the next waypoint than the safety time $T_{security}$, see Fig. \ref{sec_zone} . 

In this work, we divide our problem in 3 subproblems depending on the state of the aircraft:
\begin{enumerate}
   \item Generate base trajectories.
   \item Modify trajectories outside security zone
   \item Modify trajectories inside security zone
\end{enumerate}

The following will discuss how to deal with each of these problems in detail.

\subsection{Generating base trajectories}
When no trajectory is generated yet or the UAV finishes following the current trajectory, the next trajectory is generated from scratch, which means that does not  take in account the computation time.

For generating this trajectory, we use minimum snap polynomial-based multiwaypoint trajectory planning algorithms \cite{richter2016polynomial}\cite{burri2015real-time} due to their simplicity and computation speed. 

For generating these trajectories, a set of ordered dynamic waypoints must be provided, the order of each waypoint represents the order in which the UAV will reach each waypoint. 
When a trajectory is generated from scratch, we constraint the maximum speed and acceleration to ensure the feasibility of the trajectory generated.

For generating trajectories in this way, it is necessary to know the state of the uav when the trajectory generation process starts, for generating trajectories that start from the position of the uav. 

\subsection{Modify trajectories outside security zone}

% When we want to modify the trajectory that the aircraft is currently following $\mathbf{F}(t)$, because we want to modify the position of a set of dynamic waypoints or because a new waypoint is wanted to be added to the trajectory, we have to consider if the aircraft is inside a security zone or not.

If the aircraft is outside the security zone, the trajectory can be modified by generating a new base trajectory $\tilde{\textbf{P}}(t)$ from scratch, updating the position of the dynamic waypoints, or adding new ones.

While the new trajectory is being generated, the UAV is going to continue following the old trajectory until the new trajectory is computed and the old trajectory is replaced.
% In order to ensure the smoothness during the whole track, this new trajectory must be generated in a convenient way for generating an smooth stitching between the old trajectory and the new trajectory.
To ensure the smoothness during the whole track, this new trajectory must be generated taking into account an smooth stitching between the old trajectory and the new trajectory.

\begin{figure}[thpb]
\centering
  \includegraphics[width=0.45\textwidth]{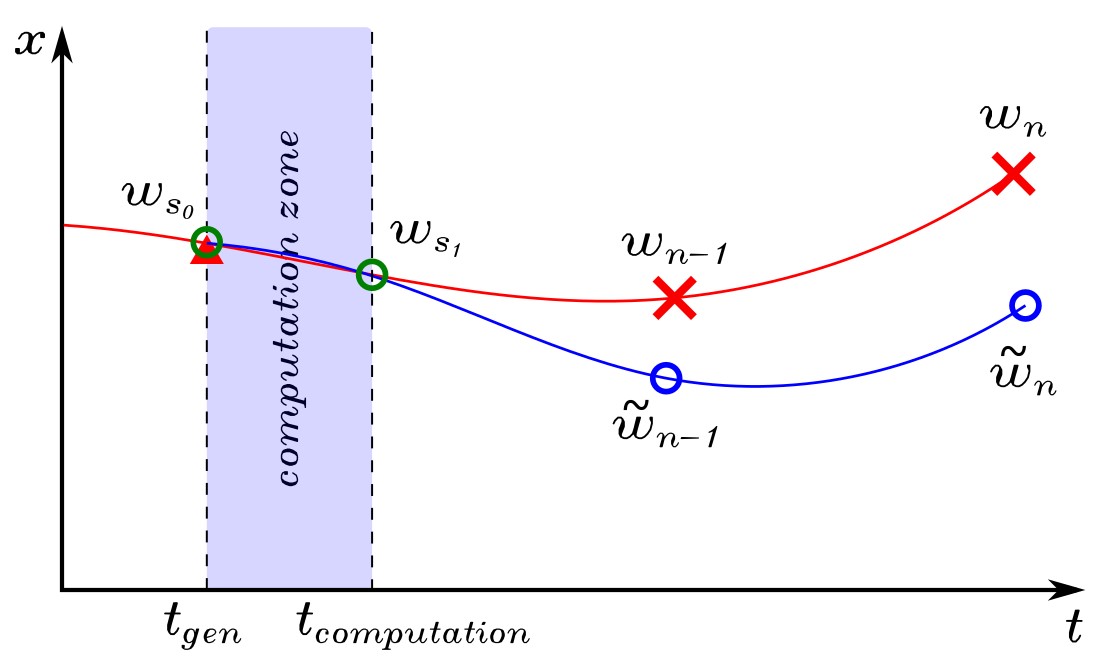}
  \caption{New base trajectory (blue) generated from another trajectory (red) using swapping waypoints (green circles) and waypoints modified (blue circles). Red crosses represent the previous localization of each waypoint and the red triangle the position of the uav when the new trajectory is calculated. }
  \label{def_stitching}
\end{figure}

If both trajectories are too different at the swapping moment, the trajectory followed by the aircraft would have a discontinuity that breaks the smoothness of the whole trajectory. To minimize this discontinuity in the trajectory swapping, the new trajectory will be computed using a set of $N_{smooth}$ waypoints (swapping waypoints) for smoothing this swapping, and the set of waypoints through which the trajectory is going to pass, see Fig. \ref{def_stitching}.

% \begin{figure}[thpb]
% \centering
%   \includegraphics[width=0.45\textwidth]{images/stitching.jpg}
%   \caption{Esquema stitching Security Zone}
%   \label{figurelabel}
% \end{figure}

This set of swapping waypoints are obtained from the old trajectory to ensure that the new base trajectory is similar to the former one during the computational time $T_{computation}(n)$:
\begin{align}
    \mathbf{w_i} &= \mathbf{F}(t_{gen} + i \cdot t_{offset})        & i = 0,..,N_{smooth} \\
    \mathbf{\dot w_i} &= \mathbf{\dot F}(t_{gen} + i \cdot t_{offset})   & i = 0,..,N_{smooth} \\
    \mathbf{\ddot w_i} &= \mathbf{\ddot F}(t_{gen} + i \cdot t_{offset})   & i = 0,..,N_{smooth}
\end{align}
where $t_{gen}$ is the time where the new trajectory starts to be generated, $t_{offset}$ represents a temporal displacement between each waypoint. To ensure that all this waypoints are between the computation time of the new trajectory, $t_{offset}=\alpha \cdot T_{computation}(n) / N_{smooth}$ with $\alpha = 1.5$ . In this work, we tried different values for $N_{smooth}$, finding that 1 and 2 are the most convenient values.

% Sometimes the swapping waypoints added can be placed after the dynamic waypoints, which cause weird trajectories that go forward and back in an erratic way. To prevent this, we filter the waypoints that are between the swapping waypoints, because the trajectory is going to pass through them 

Finally, when the new trajectory is computed, it replaces the previous one from that point in time.
% \subsection{USER operations:}
% \begin{itemize}
%     \item Set Waypoints
%     \item Append Waypoint
%     \item Modify Waypoint
% \end{itemize}

\subsection{Modify trajectories inside security zone}
When the aircraft is inside the security zone means that it is not able to recompute a base trajectory in a safe way, without having the possibility to correct the trajectory in a robust way. In this situation, we have to use faster but sub-optimal approaches in exchange for being able to make these modifications. 

In order to be able to perform small modifications in the trajectory, we propose to do local modifications in the trajectory near the modified waypoint so the trajectory will pass through them in a smooth and agile way. For doing these modifications, we use LGMs, which computation time is more that two orders of magnitude lower than generating a base trajectory.

\begin{figure}[thpb]
\centering
  \includegraphics[width=0.45\textwidth]{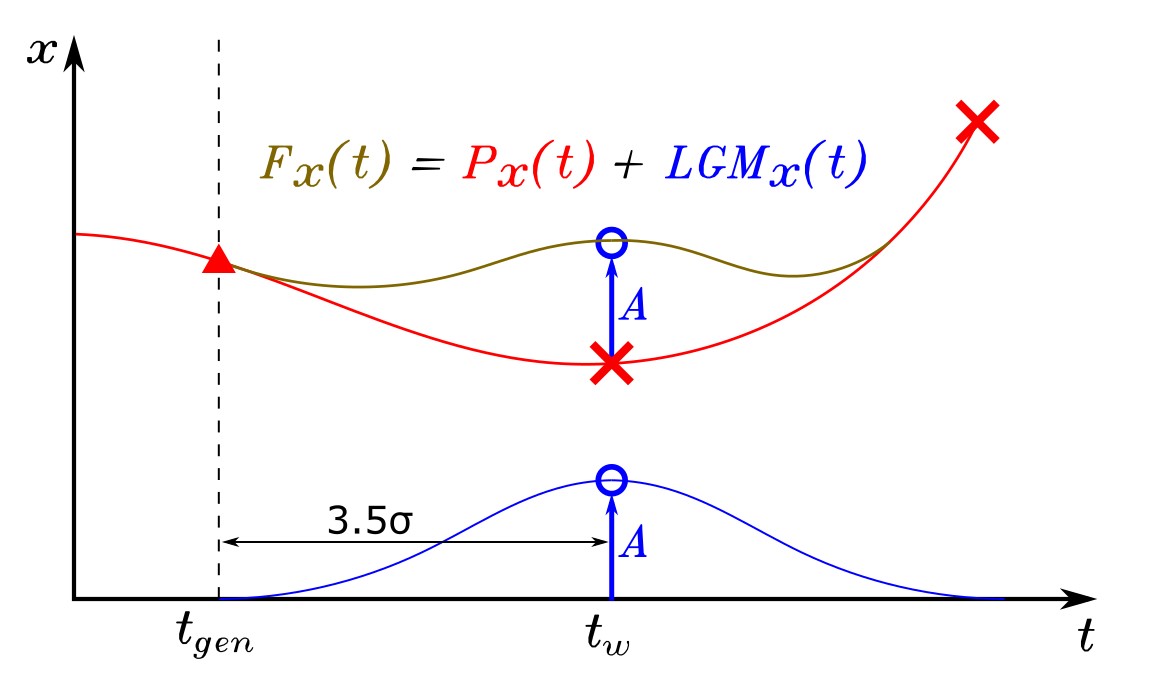}
  \caption {Representation 1-dimensional of a trajectory generated (green) with a base trajectory (red) and a Local Gaussian Modification (blue) of the waypoint $w$ with a displacement $A$ in x axis. The red triangle, represents the position of the UAV when the modification is done, red crosses represent the original position of two waypoints and the blue circle the modification of the waypoint $w$.}
  \label{def_lgm}
\end{figure}

Each LGM has 3 constants related to it: $\textbf{A}$ , $\sigma$ and $\mu$. To obtain the desired behavior, each constant must be computed before adding the modificator to the trajectory. 

Given a trajectory $\mathbf{F}(t)$ consisting on $N$ waypoints, each waypoint $w_i$ has a position $\mathbf{x}_i$ and a time in the trajectory $t_{w_i}$, which means that $\mathbf{F}(t_{w_i}) = \mathbf{w_i}$. When a modification $\tilde{\mathbf{w_i}}$ occurs in a time $t_{mod}$, a new modifier $\mathbf{LGM}_i(t)$ is generated, see Fig. \ref{def_lgm}. Each constant of the modifiers are computed in this way:

\begin{align}
    \mathbf{A} &=  \begin{bmatrix} 
    A_x\\
    A_y\\
    A_z
    \end{bmatrix} = \mathbf{\tilde{w_i}} - \mathbf{w_i}  =
    \begin{bmatrix}
    \tilde{w_i}_x - {w_i}_x\\
    \tilde{w_i}_y - {w_i}_y\\
    \tilde{w_i}_z - {w_i}_z
    \end{bmatrix}  \\
    \mu &=  t_{w_i} \\
    \sigma &= \frac{|t_{mod}- \mu|}{3.5}
\end{align}
Where $\mathbf{A}$ represents the amplitude of the modification, $\mu$ represents the temporal position of the waypoint in the current trajectory and $\sigma$ the variance of the Gaussian.
These parameters are chosen in such way that $\mathbf{LGM}(t_{mod}) \approx \mathbf{0}$, with this we guarantee an smooth change in the trajectory when each modification is done. Due to the properties of the Gaussian function, the 99.98\% of the contribution of $\mathbf{LGM}(t)$ is between $t \in [-3.5\sigma < t-\mu < 3.5\sigma] $, so choosing sigma in this way let us consider that adding this modifiers to $\mathbf{F}(t)$ maintain the conditions of continuity and derivability of the trajectory. 

The low computational cost of generating and evaluating LGMs allows us to append multiples of them in the same waypoint, obtaining a high reactiveness to the trajectory changes while maintaining the smoothness condition over the trajectory.

\subsection{Implementation details.}

The implementation of this dynamic trajectory generator has been done in C++ and it is public accessible\footnote{https://github.com/miferco97/dynamic\_trajectory\_generator}.  We use a library developed at the ETH Zurich University, for generating polynomial trajectories \footnote{https://github.com/ethz-asl/mav\_trajectory\_generation}  based on the work of Ritcher et al. \cite{richter2016polynomial}

% \color{-red!50}

% In order to ensure that we can precisely follow the polynomial trajectories we intend to generate, we utilize the property of differential flatness for the standard quadrotor equations of motion:
% mr¨ = mgzW − f zB (1)
% ω˙ = J
% −1
% [−ω ×Jω +M] (2)
% Differential flatness of this model was demonstrated in [19]. Here, r is the position vector of the vehicle in a global coordinate frame, ω is the angular velocity vector in the body-fixed coordinate frame and f and M are the net thrust and moments in the body-fixed coordinate frame. J and m are the inertial tensor and mass of the quadrotor. zB is the unit vector aligned with the axis of the four rotors and indicates the direction of thrust, while zW is the unit vector expressing the direction of gravity. There exists a simple mapping from f and M to the four desired motor speeds 

% Due to the need of generating trajectories for the controller, a polynomial trajectory generator has been used. The trajectories are generated on a set of waypoints and are optimal in acceleration, which guarantees smoothness in the actuator commands. Moreover, the trajectories generated are constrained by maximum speed vmax and maximum acceleration amax parameters. 

% \newpage
% \newpage
%%%%%%%%%%%%%%%%%%%%%%%%%%%%%%%%%%%%%%%%%%%%%%%%%%%%%%%%%%%%%%%%%%%%%%%%%%%%%%%%
\section{EXPERIMENTS AND RESULTS}

\subsection{Experimental setup}
For the simulation experiments, a laptop with Ubuntu 20.04 and an Intel Processor i7-10870H @2.2GHz has been used. In the real flight we use a Nvidia Jetson Xavier NX with a CPU NVIDIA Carmel ARM®v8.2 of 64 bits. We perform profiling tests in both computers.

All Flight experiments have been performed using a ROS2 version of the recent framework for autonomous drone racing based on Aerostack 5.0 \cite{imav2021-5:m_fernandezcortizas_et_al}.

\subsection{Local Gaussian Modification tests}

The first experiment consists on testing the performance of the dynamic trajectories generated only using LGMs for updating the trajectory. 

\begin{figure}[thpb]
\centering
  \includegraphics[width=0.49\textwidth]{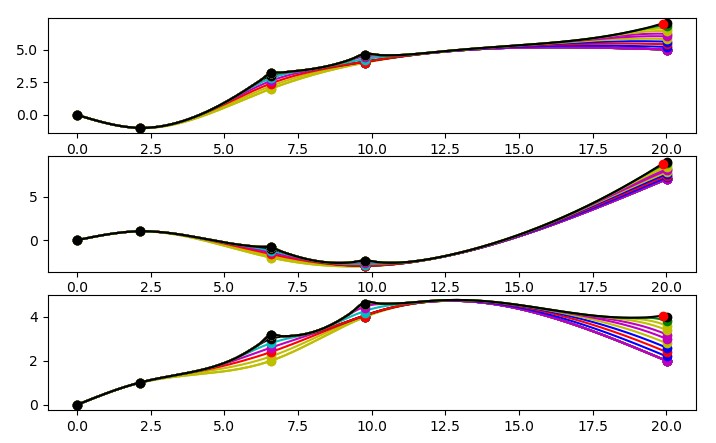}
  \caption{Axis by axis (x, y and z axis, corresponds with top, middle and bottom subfigure respectively) plot of the dynamic trajectory position references generated using only LGMs to modify the waypoint positions for different modifications of some waypoints. The vertical axis represents the position on each axis in meters, while the horizontal represents the time in seconds.}
  \label{only_gaussians}
\end{figure}

\begin{figure}[thpb]
\centering
  \includegraphics[width=0.49\textwidth]{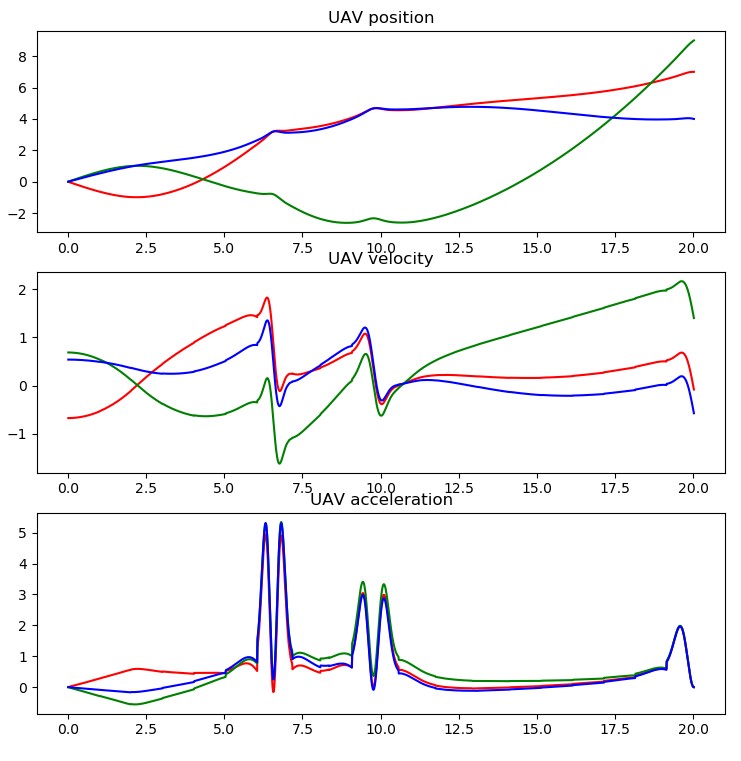}
  \caption{Position, velocity and acceleration references generated by the trajectory generation during the following of the dynamic trajectory shown above in Fig. \ref{only_gaussians} .}
  \label{references}
\end{figure}

For this experiment, we generate a trajectory with 5 waypoints and as long as the UAV goes to it, some waypoints increase his distance to the former trajectory. Fig. \ref{only_gaussians} shows the different modifications realized over the main trajectory, Fig. \ref{references} shows the references obtained by the quadrotor during the whole track.

Some profiling tests about the time spent in computing base trajectories (TABLE \ref{profile_base}) and the time spent generating and evaluating LGMs, have been done both in the high end computer and the onboard computer.

\begin{table}[htb]
\centering
\begin{tabular}{l|c|c}
    $\empty$ & \textbf{Laptop computer} & \textbf{Jetson Xavier NX} \\
%  	\midrule
    %\hline
	6 points  & $13.71  \pm 0.14  \; ms$ &	$51.47	 \pm 1.57  \; ms$\\
	14 points & $88.93	 \pm 3.11  \; ms$ &	$321.07	 \pm 22.11 \; ms$\\
	26 points & $309.08 \pm 10.20 \; ms$ &	$1027.76 \pm 19.36 \; ms$\\
\end{tabular}
	\caption{Time spent generating base polynomial trajectories for different number of points, both in a computer and in a on board computer.}
	\label{profile_base}
\end{table}

\begin{table}[htb]
\centering
\begin{tabular}{l|c|c}
    $\empty$ & \textbf{Laptop computer} & \textbf{Jetson Xavier NX} \\
%  	\midrule
    %\hline
	1 LGMs     & $52.9  \pm 8  \; ns$ &	$207	 \pm 54  \; ns$\\
	8 LGMs     & $223 \pm 34  \; ns$ &	$1123    \pm  112 \; ns$\\
	64 LGMs    & $1544 \pm 303 \; ns$ &	$7694	 \pm 442 \; ns$\\
	512 LGMs   & $12451  \pm 504 \; ns$ &	$61374	 \pm  829\; ns$\\
\end{tabular}
	\caption{Time spent generating and evaluating the sum of multiple LGMs acting on the same trajectory.}
	\label{profile_lgm}
\end{table}

\subsection{Simulation flights}

For this experiment, 4 dynamic gates have been placed in the circuit. These gates moved from side to side up to 1 meter and at a constant speed of 0.1m/s. All simulation has been done in the gazebo simulator, using ROS2 for the communication between modules. 

To study the effect of the computation time of the system on the generation and modification of the trajectories, 3 rounds of experiments were carried out: one with the computation time taken by the computer to generate a trajectory, and then increased by 0.5s and 1 second, see TABLES \ref{tab:sim:ct0.0} \ref{tab:sim:ct0.5} \ref{tab:sim:ct1.0}.
\begin{table*}[]
    \centering
    \begin{tabular}{|c|c|c|c|c|c|c|c|c|}
% Trajectory 
\hline
speed limit &	Max speed	&Mean speed	&Max speed	&mean speed	&time elapsed	&success \\
\hline
5	 &5.15      &1.54       &5.25       &2.23       &48.5       &0.88   \\					
10	&8.59	&2.34	&9.07	&2.94	&35.96	&0.85 \\
15 &13.05        &2.78       &11.41       &3.55       &31.1       & 0.9  \\ 
20	&10.28	&1.9	&11.19	&2.45	&44.7	&0.88 \\
\hline
\end{tabular}
    \caption{Results for simulated flight with computing time}
    \label{tab:sim:ct0.0}
\end{table*}

\begin{table*}[]
    \centering
    \begin{tabular}{|c|c|c|c|c|c|c|c|c|}
% Trajectory 
\hline
speed limit &	Max speed	&Mean speed	&Max speed	&mean speed	&time elapsed	&success \\
\hline
5	 &4.47      &1.87       &5.10       &2.39       &42.30       &0.90   \\			
10	&9.50	&3.53	&9.48	&3.61	&29.54	&0.90 \\
15 &13.63        &3.27       &13.80       &3.35       &31.25       & 0.95  \\ 
20	&15.64	&3.58	&16.01	&3.70	&29.95	&0.75 \\
\hline
\end{tabular}
    \caption{Results for simulated flight with computing time + 0.5s}
    \label{tab:sim:ct0.5}
\end{table*}

\begin{table*}[]
    \centering
    \begin{tabular}{|c|c|c|c|c|c|c|c|c|}
% Trajectory 
\hline
speed limit &	Max speed	&Mean speed	&Max speed	&mean speed	&time elapsed	&success \\
\hline
5	 &4.71      &2.25       &4.7       &2.29       &45.9 0      &0.92   \\					
10	&9.63	&3.33	&11.08	&3.46	&31.1	&0.85 \\
15 &13.61        &4.77       &13.82       &4.87       &21.80       & 0.90  \\ 
20	&15.58	&4.06	&16.72	&3.24	&27.47	&0.65 \\
\hline
\end{tabular}
    \caption{Results for simulated flight with computing time + 1.0s}
    \label{tab:sim:ct1.0}
\end{table*}

\begin{tabular}{@{}l|c|c@{}}
    $\empty$ & Computer & Jetson Xavier NX \\
%  	\midrule
    %\hline
	6 points  & $13.71  \pm 0.14  \; ms$ &	$51.47	 \pm 1.57  \; ms$\\
	14 points & $88.93	 \pm 3.11  \; ms$ &	$321.07	 \pm 22.11 \; ms$\\
	26 points & $309.08 \pm 10.20 \; ms$ &	$1027.76 \pm 19.36 \; ms$\\
\end{tabular}

\subsection{Real flight experiment.}

\begin{figure}[thpb]
\centering
  \includegraphics[width=0.40\textwidth]{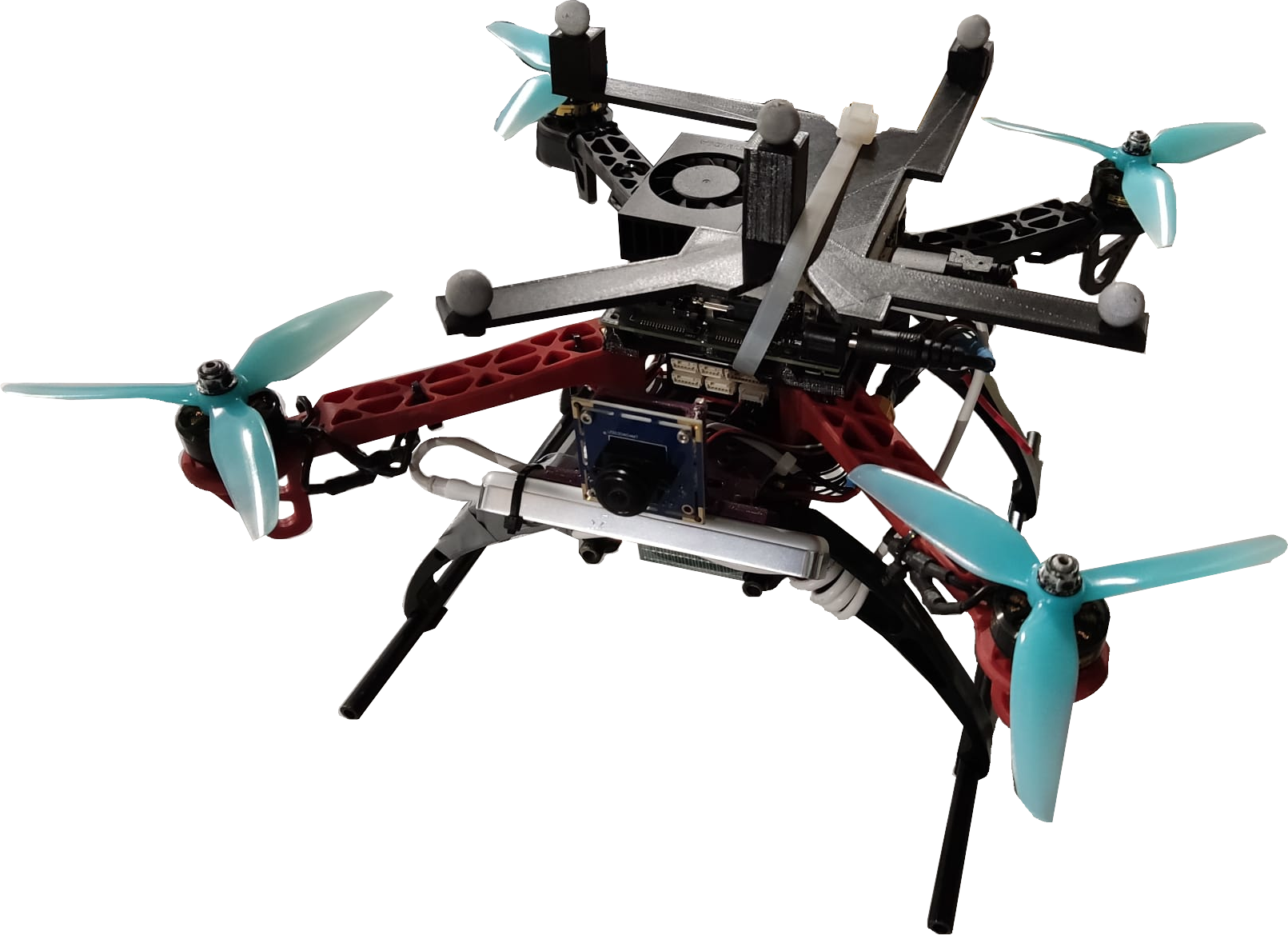}
  \caption{Quadrotor used for real flight experiments}
  \label{fig:drone}
\end{figure}

The aerial platform used for the real experiments was a custom quadrotor based on the DJI F330 frame, shown in Fig. ~\ref{fig:custom_platform}. This platform was equipped with a Pixhawk 4 mini as the aircraft autopilot, an Intel Realsense T265 Tracking Module used for state estimation, and an USB fish-eye camera for gate detection. Additionally, the aerial platform was equipped with a Single Board Computer (SBC) NVIDIA Jetson Xavier NX with an 6-core ARM v8.2. Fig. \ref{fig:drone}.

For validating the approach in a real environment, an additional experiment in real was done. The experiment consisted of passing through a small circuit of two gates each lap faster, beginning with a maximum speed of 0.5m/s and increasing it 0.5 m/s each lap until 3.5m/s, see Fig. \ref{real}. 

\begin{figure}[thpb]
\centering
  \includegraphics[width=0.49\textwidth]{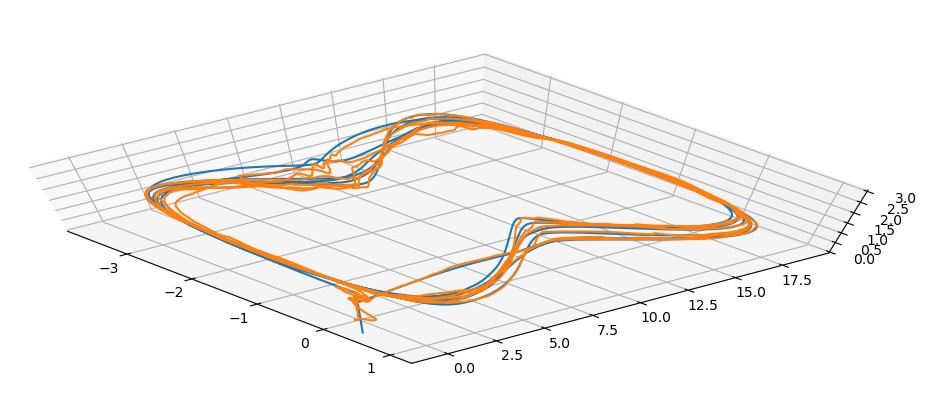}
  \caption{Trajectory generated during the real flight (blue), and the trajectory followed by the UAV (orange)}
  \label{real}
\end{figure}

\begin{figure}[thpb]
\centering
  \includegraphics[width=0.49\textwidth]{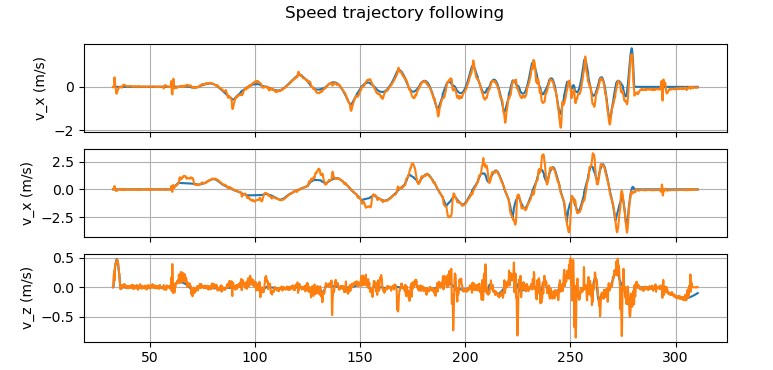}
  \caption{Plot per axis of the Trajectory Speed generated during the real flight (blue), and the trajectory speed followed by the UAV (orange)}
  \label{real}
\end{figure}

%%%%%%%%%%%%%%%%%%%%%%%%%%%%%%%%%%%%%%%%%%%%%%%%%%%%%%%%%%%%%%%%%%%%%%%%%%%%%%%%
\color{black}
\section{DISCUSSION}
In the first testbed, we can see that the trajectory modificated with the LGMs reaches all the waypoints in a smooth way, generating smooth references for position, velocity and acceleration, as we can see in Fig. \ref{only_gaussians} and Fig. \ref{references}. Moreover, for the profiling tests we observe that modifying a trajectory with LGMs is more than two orders of magnitude faster than generating a new polynomial trajectory for the usual number of LGMs in a trajectory (between 8 to 64 LGMs).

In the simulated experiments, it has been possible to prove that with a very small calculation time, like that of a high capacity computer, it is capable of constantly regenerating the trajectory each time it detects a change in the position of one of the waypoints. Moreover, when increasing the calculation time, trying to make it similar to the time of computers with lower computation capacity, it is no longer able to generate trajectories and has to make modifications to them using LGMs. This effect increases considerably with speed. With all this, speeds in excess of 14 m/s are achieved by consistently passing through the moving doors.

In the last experiment, we tested our approach in the real world, being able to achieve peak speeds up to 4 m/s, with an average speed of 1 m/s during the experiment. We were not able to fly faster because of the uncertainties in the perception and state estimation module, which added noise to the trajectory following leading to crash.

%%%%%%%%%%%%%%%%%%%%%%%%%%%%%%%%%%%%%%%%%%%%%%%%%%%%%%%%%%%%%%%%%%%%%%%%%%%%%%%%
\color{black}
\section{CONCLUSIONS}

In this work, a novel method to modify base trajectories, whose calculation is computationally expensive, when the reaction time and computing resources are limited, was presented, being able to achieve high speeds up to 16m/s combined with a polynomial trajectory generator in a simulated dynamic environment, being robust to changes in the base trajectory computation time. In real flight, the lack of accuracy in gate estimation and state estimation results in poorer system performance. This work can be really useful for very low computational power devices as a main way to generate dynamic trajectories.

% As our algorithm is a first version (under development) there are many aspect that can be improved
Although the use of Local Gaussian Modifier as a modifier function has benefits such as the increase in the reactivity of the trajectory generator in onboard computing circumstances, they also show some shortcomings for this technique. For example, the unbounded influence of Gaussian function can cause nearby points to add effects, causing the trajectory to be modified beyond the desired point. In addition, modifying the generated trajectory to meet the physical limits of the platform may cause the speed and acceleration limits to be exceeded. These problems can be faced looking for other modifier functions that take into account these limitations and modify the trajectory in a different way.

Another possibility to study can be to scale this philosophy to be used with very long time computation trajectory generator algorithms, such as CPC \cite{foehn2021time}, combined with a polynomial trajectory generator as the local modifier, to be able to exploit the optimally of the CPC with the low computational cost of the polynomial trajectory.

% \tb{las gaussianas pueden hacer que te salgas de los limites (v,a). buscar otras funciones, espaciar el tiempo, externder el tiempo de la trayectoria principal para asegurar que se cumnplan las limitaciones v and a del drone}

%                                   % the textheight of the last page by a suitable amount.
%                                   % This command does not take effect until the next page
%                                   % so it should come on the page before the last. Make
%                                   % sure that you do not shorten the textheight too much.
%%%%%%%%%%%%%%%%%%%%%%%%%%%%%%%%%%%%%%%%%%%%%%%%%%%%%%%%%%%%%%%%%%%%%%%%%%%%%%%%
\color{black}

%%%%%%%%%%%%%%%%%%%%%%%%%%%%%%%%%%%%%%%%%%%%%%%%%%%%%%%%%%%%%%%%%%%%%%%%%%%%%%%%
\color{black}
\section*{ACKNOWLEDGMENT}
This work has been supported by the project COMCISE RTI2018-100847-B-C21, funded by the Spanish Ministry of Science, Innovation and Universities (MCIU/AEI/FEDER, UE) and the project “COPILOT: Control, Supervisión y Operación Optimizada de Plantas Fotovoltaicas mediante Integración Sinérgica de Drones, IoT y Tecnologías Avanzadas de Comunicaciones” Ref: Y2020/EMT6368 Funded by Madrid Government under the R\&D Sinergic Projects Program. 

% The preferred spelling of the word *acknowledgment* in America is without an *e* after the *g*. Avoid the stilted expression, *One of us (R. B. G.) thanks . . .*  Instead, try *R. B. G. thanks*. Put sponsor acknowledgments in the unnumbered footnote on the first page.
%%%%%%%%%%%%%%%%%%%%%%%%%%%%%%%%%%%%%%%%%%%%%%%%%%%%%%%%%%%%%%%%%%%%%%%%%%%%%%%%
% \textcolor{blue}{REVISAR FORMATO BIBLIOGRAFIA}
\bibliographystyle{unsrt}
\bibliography{bibliography}
% \input{bibliography}
% \cite*
%%%%%%%%%%%%%%%%%%%%%%%%%%%%%%%%%%%%%%%%%%%%%%%%%%%%%%%%%%%%%%%%%%%%%%%%%%%%%%%%
%\input{style_guide}
\end{document}